\documentclass[conference]{IEEEtran}

  	\usepackage[pdftex]{graphicx}
  	\graphicspath{{../pdf/}{../jpeg/}}
	\DeclareGraphicsExtensions{.pdf,.jpeg,.png}

	\usepackage[cmex10]{amsmath}
	\usepackage{mathabx}
	\usepackage{algorithmic}
	\usepackage{array}
	\usepackage{mdwmath}
	\usepackage{mdwtab}
	\usepackage{eqparbox}
	\usepackage{url}
	\usepackage{multirow}
    \usepackage{subcaption} 
\begin{document}

\title{\LARGE NukeBERT:  A Pre-trained language model for Low Resource Nuclear Domain}

% \author{\authorblockN{Leave Author List blank for your IMS2013 Summary (initial) submission.\\ IMS2013 will be rigorously enforcing the new double-blind reviewing requirements.}
% \authorblockA{\authorrefmark{1}Leave Affiliation List blank for your Summary (initial) submission}}

 \author{\authorblockN{Ayush Jain\authorrefmark{1}, Dr. N.M. Meenachi\authorrefmark{2}, Dr. B. Venkatraman\authorrefmark{2}}
 \authorblockA{\authorrefmark{1}BITS PILANI, Pilani, India, \authorrefmark{2}IGCAR, Kalpakkam, India}
 \authorblockA{\authorrefmark{1}ayushjain1144@gmail.com \authorrefmark{2}meenachi@igcar.gov.in, \authorrefmark{2}bvenkat@igcar.gov.in}}

\maketitle

\begin{abstract}
Significant advances have been made in recent years on Natural Language Processing with machines surpassing human performance in many tasks, including but not limited to Question Answering. The majority of deep learning methods for Question Answering targets domains with large datasets and highly matured literature. The area of Nuclear and Atomic energy has largely remained unexplored in exploiting available unannotated data for driving industry viable applications. To tackle this issue of lack of quality dataset, this paper intriduces two datasets: NText, a eight million words dataset extracted and preprocessed from nuclear research papers and thesis; and NQuAD, a Nuclear Question Answering Dataset, which contains 700+ nuclear Question Answer pairs developed and verified by expert nuclear researchers. This paper further propose a data efficient technique based on BERT, which improves performance significantly as compared to original BERT baseline on above datasets. Both the datasets, code and pretrained weights will be made publically available, which would hopefully attract more research attraction towards the nuclear domain. 
\end{abstract}

% This paper introduces a new nuclear dataset, created from the 7000 research papers on nuclear domain. This paper contributes to research in understanding nuclear domain knowledge which is then evaluated on Nuclear Question Answering Dataset (NQuAD) created by nuclear domain experts as part of this research. NQuAD contains 612 questions developed on 181 paragraphs randomly selected from the IGCAR research paper corpus. In this paper, the Nuclear Bidirectional Encoder Representational Transformers (NukeBERT) is proposed, which incorporates a novel technique for building BERT vocabulary to make it suitable for tasks with less training data. The experiments evaluated on NQuAD revealed that NukeBERT was able to outperform BERT significantly, thus validating the adopted methodology. Training NukeBERT is computationally expensive and hence we will be open-sourcing the NukeBERT pretrained weights and NQuAD for fostering further research work in the nuclear domain.

\IEEEoverridecommandlockouts

\IEEEpeerreviewmaketitle

% ===================
% # I. Introduction #
% ===================

\section{Introduction}
The nuclear industry presents a viable solution to end energy crisis. Advancements in machine and deep learning is playing a
significant role in enhancing the progress in the medical domain (\cite{1}, \cite{2}, \cite{3}) and a lot of groundwork is already done in  generation of various datasets and
pretrained models, thus reducing the barrier for future research.  Similar advancements needs to happen in the nuclear
field. A survey indicated that limited effort has gone into the domains of power plants and atomic energy\cite{4}. This research aims to develop quality datasets and explore the applicability of BERT (Bidirectional Encoder
Representations from Transformers)\cite{5} in the nuclear field, which lacks high quality
data.  

\begin{figure}[ht!] %!t
\centering
\includegraphics[width=3.4in]{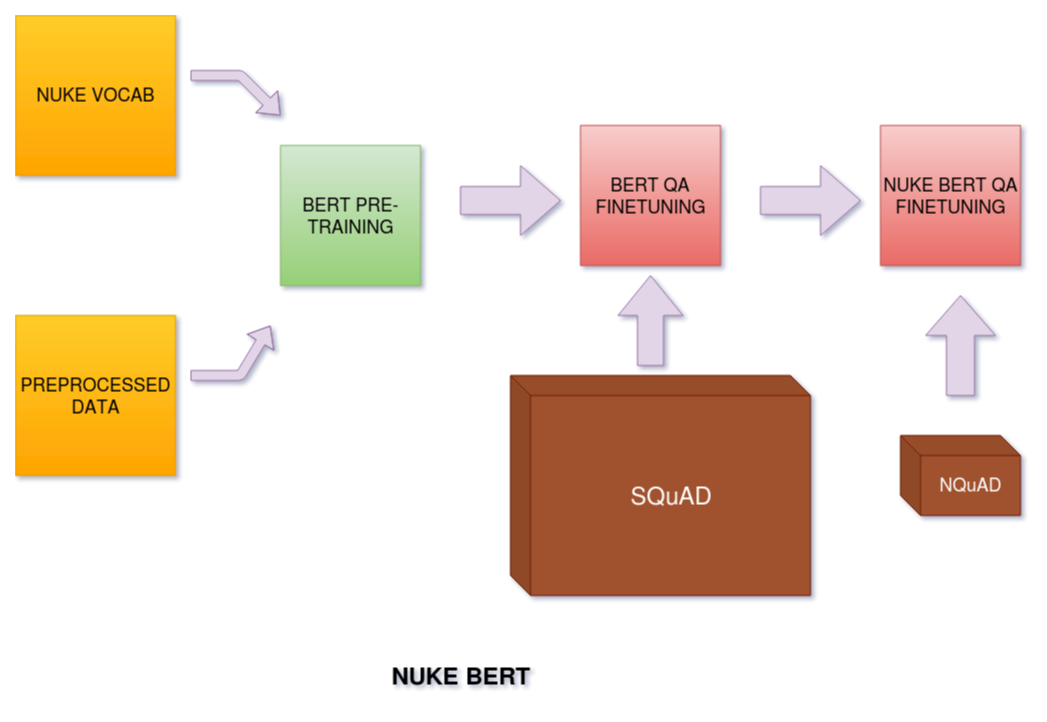}
\caption{ \textbf{NukeBERT Pipeline}: Using a vocabulary having nuclear jargons, the preprocessed data is pre-trained on a corpus of nuclear text, which is then further finetuned on SQuAD and NQuAD.}
\label{Courant_2}
\end{figure}

On this ground, a model is proposed in this research which will be referred to as Nuclear Bidirectional Encoder Representations from Transformers for language understanding (NukeBERT) (Figure - \ref{Courant_2}). NukeBERT is a contextualized nuclear word embedding, based on BERT, which can be fine-tuned further on numerous downstream tasks including Question Answering, Named-entity recognition and Sentence segmentation. BERT is trained on wikipedia corpus, which is strikingly different and lacking in nuclear terminologies. Further, BERT requires a huge corpus for generating word embeddings, which is difficult to build in a low resource nuclear domain. To incorporate the nuclear jargons, we developed a custom dataset, “NText”, for training BERT. The model is then evaluated for its performance on Question Answering task. Due to lack of question-answering dataset in the nuclear domain, we developed a new dataset, called NQuAD (Nuclear Question Answering Dataset), prepared with the help of nuclear domain experts at Indira Gandhi Center for Atomic Research (IGCAR) as part of this research.

With the same configuration as used by original BERT model for fine-tuning, NukeBERT is fine-tuned on Stanford Question Answering Dataset (SQuAD)\cite{6} and then on NQuAD train set. On testing, the NukeBERT outperformed the original BERT model significantly.

Our contributions are as follows: 1) We introduce a novel nuclear natural language datasel called NText, which can be further used for numerous language modelling tasks.
2) We further introduce the first nuclear question answering dataset, which has expert curated 700+ question answer pairs which can be used to train and test various NLP models for nuclear domain. 3) We further propose a novel state-of-the-art technique which is able to achieve superior performance while using orders of magnitude less data. 

For fostering future research and industrial applications, we will be open-sourcing NText, NQuAD and pretrained weights.

% =======================================================
% # II. Related Work #
% =======================================================

\section{Related Work}
Word2Vec \cite{word2vec} pioneered the semi-supervised approach to read massive corpora and generate meaningful word embeddings. However, vanilla Word2Vec suffers massively from polysemy, i.e. words with different meaning in different contexts and inability to effectively deal with out of vocabulary words (\cite{7}, \cite{8})

The advent of ELMO (Embeddings from Language Models) \cite{9}, GPT (Generative Pre-Training Model) \cite{10} and BERT demonstrated the benefit of unsupervised language modelling pre-training on large corpus for various sophisticated downstream tasks. While GPT and ELMO required task specific architectures, BERT showed excellent results on various NLP tasks and benchmarks with minimum need of architectural modifications. BERT was trained on 3.3 Billion words dataset to generate BERT embeddings, which were then fine-tuned to various downstream tasks. With minimal architecture modification, BERT was able to achieve a state of the art in 11 natural language processing tasks.

SciBERT, a BERT based approach proposed for scientific data, was trained on 1.14 Million papers from semantic scholar having 3.17 Billion tokens. It was able to improve the model accuracy on several scientific and biomedical datasets than BERT \cite{11}. BioBERT, proposed for biological domain, is trained on 18 Billion words dataset of medical corpus and 3 Billion words BERT dataset. BioBERT could significantly improve the state-of-the-art performance on various medical datasets. \cite{12}.

In contrast to BERT, SciBERT and BioBERT, nuclear domain suffers from lack of large, high quality datasets. After collecting data from nuclear research papers. a 8 Million words nuclear dataset was generated, which is two magnitudes lower than dataset by the existing BERT based approaches. Hence a novel approach to optimize the formation of Nuclear Vocab (NukeVOCAB) and pre-training to achieve better results than BERT is developed. Unlike the medical domain, the nuclear domain do not have any gold training dataset to gauge the performance of this newly developed model. Hence a Nuclear Question Answering Dataset (NQuAD) is developed with the effort of Nuclear scientists. The domain experts have evaluated the system for its performance. The NukeBERT methodology is explained in the following section.

% =============================================
% # III. Corpus Preparation #
% =============================================
\section{Methodology}

\subsection{NText Preparation}

This research work proposes NText, which is a Nuclear Textual dataset containing textual data related to nuclear domain. For the dataset preparation, 7000 internal reports, thesis and research papers in the PDF format were taken from the Indira Gandhi Centre for Atomic Research (IGCAR). The sizes of the reports ranged from a couple of pages to a few thousand pages. A substantial portion of nuclear corpus consisted of very old reports, some of which were stored as scanned copies. The reports primarily dealt with the nuclear domain, many of them explicitly dealing with Fast Breeder Reactor(FBR). The raw corpus needed extensive cleaning and preprocessing to convert it into pre-training corpus suitable for BERT. The next section will discuss the preprocessing pipeline in detail.

\begin{figure}[ht!] %!t
\centering
\includegraphics[width=3in]{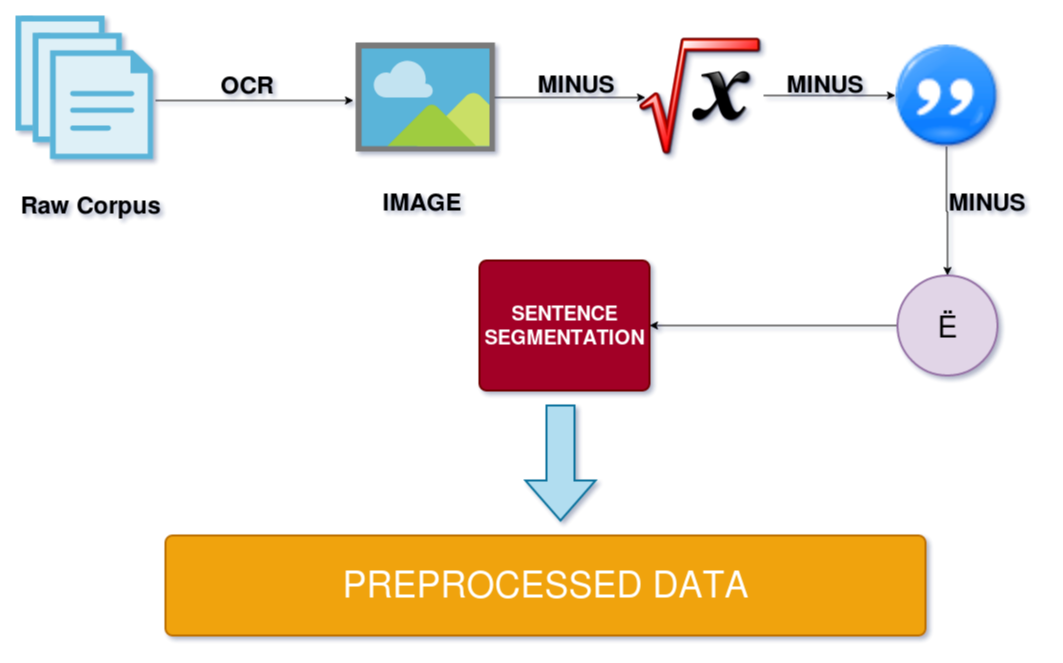}
\caption{Flowchart of steps for preparation of preprocessed data from raw corpus of 7000 research papers.}
\label{Courant_3}
\end{figure}

\subsection{PREPROCESSING}

The detailed pipeline for preprocessing is shown in Figure - \ref{Courant_3}. Since the dataset had a lot of scanned copies which are unreadable for PDF parsing libraries, it was decided to use Optical Character Recongnition (OCR) for extracting the text.  Another benefit of using OCR is that it avoids random unicode errors which typically occur while parsing PDFs and would have been unavoidable considering the diversity of the PDFs in the corpus used. Tesseract-Optical Character Recognition (OCR) \cite{13} is used to read the text from the images. The OCR is not accurate and generates specific errors like reading 'o' as '0', 'I' as '1'. However, according to Namsyl et al. \cite{14}, these errors are infact a useful data augmentation technique. Further, it is not a significant issue since BERT tokenizer will be tokenizing the errorneous word by splitting it at the wrongly comprehended alphabet and hence would still be able to understand a meaningful representation of the word.

OCR is computationally expensive, and its accuracy and processing time highly depends upon the quality of the image. Our corpus has varied quality PDFs, with some being very old, handwritten and misaligned. It took approximately four days of continuous running on a single Google Cloud K80 Tesla GPU. The average time taken for OCR and further preprocessing (described below) was around 20 seconds/page. In general, there are many mathematical formulas in the research papers and thesis. We devised several regular expression patterns to remove them from the corpus. Further, there were a lot of references, both in-text and at the references at the end of the main text of the papers. For removing them, we designed regex patterns for common styles of in-text citations used in the literature. We noticed that last 2-3 pages generally contain references, and hence we removed last two pages from each corpus as well.

\begin{figure}[ht!] %!t
\centering
\includegraphics[width=3in]{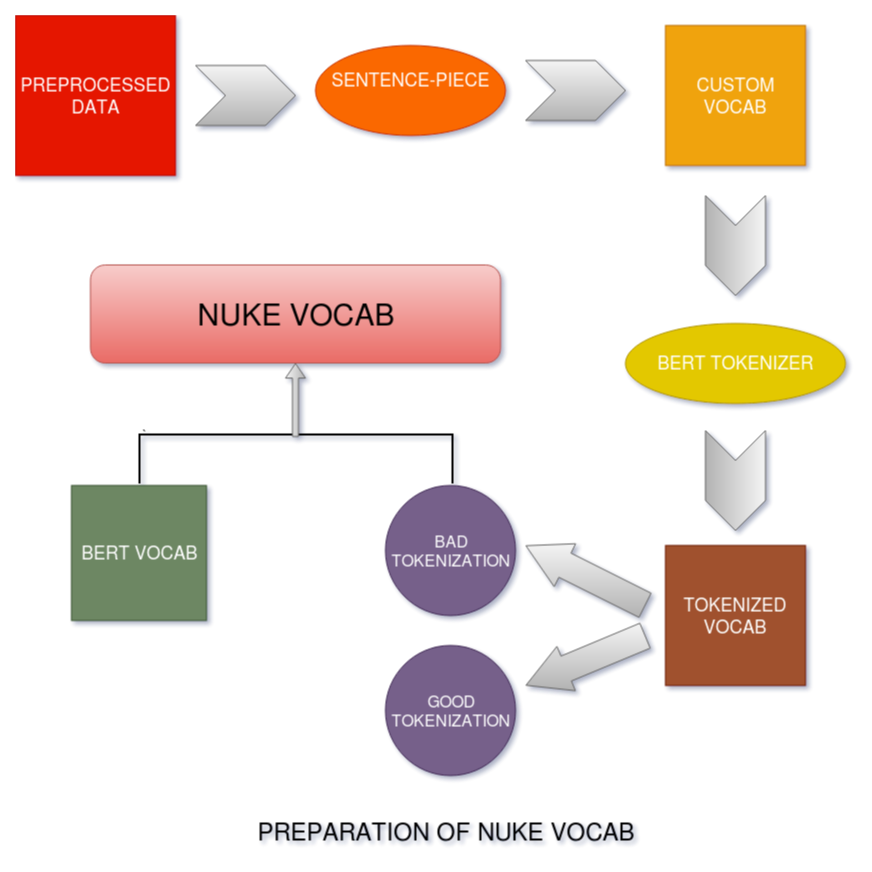}
\caption{Flowchart for preparation of NukeVocab. Notice we only concatenate “Bad Tokenization” with BERT Vocabulary.}
\label{Courant_4}
\end{figure}

Since the original BERT model is trained on general English corpus, having non-english text would not be suitable for training purposes. The raw corpus had some research papers and thesis in German and French. There are some libraries like NLTK which can approximately identify whether a word is a non-english word. However, in our experiments, we found that it was not suitable as some of the nuclear jargons were also falsely marked as non-english word. We noticed that non-english texts in the raw corpus typically contained characters like ë, è, é. Since ASCII representation only contains numbers and English alphabets, we ignored those lines which included these non ascii characters and thus removed about 20000 non-english lines from the corpus. Further, we also manually removed the non-english texts from the raw corpus.

The input for BERT pre-training is a text file with one sentence per line. An empty line delimits the documents. We used the gensim library for segmenting the text into sentences, which finally makes the input text file ready for pre-training and other tasks.
After all the cleaning and preprocessing, we obtained NText, a Nuclear textual corpus consisting of about 8 Million words.

\subsection{NukeVOCAB}

BERT makes use of a technique referred as WordPiece in their paper, to build a vocabulary on their general words corpus. BERT uses a vocabulary of thirty two thousand words. Since the nuclear field contains a lot of technical words, typically missing from the general texts,  it is essential to have a vocabulary tailored to the nuclear domain. Hence we build a Nuclear Vocabulary of thirty thousand words which will be referred  as NukeVocab (Nuclear Vocabulary) in further discussions. The procedure for constructing NukeVocab is summarized in Figure - \ref{Courant_4}. First, Sentence Piece library released by tensor2tensor is used to generate a list of thirty thousand words similar to BERT vocabulary from NText data. We will refer it as Custom-Vocab (Figure - \ref{c_vocab}), which will be further modified to obtain NukeVocab. Custom-vocab is essentially a list of essential and frequently occuring words from NText corpus. The preprocessed NText was given as input which generated a output vocabulary of essential words and subwords in an unsupervised fashion. We modified the output of sentence piece to make it similar to BERT vocabulary.

% \begin{figure}
%   \begin{subfigure}[b]{1.4in}
%     \includegraphics[width=1.4in]{fig4.png}
%     \caption{BERT tokenization of Custom-Vocab}
%     \label{fig1:1}
%   \end{subfigure}
%   %
%   \begin{subfigure}[b]{1.4in}
%     \includegraphics[width=1.4in]{Fig5.png}
%     \caption{Custom Vocab generated using sentence piece}
%     \label{fig1:2}
%   \end{subfigure}
%  \caption{Custom Vocabulary}
%   \label{c_vocab}
% \end{figure}

\begin{table}
\centering
\large
\begin{tabular}{|c|c|} 
 \hline
 \textbf{Custom Vocabulary} & \textbf{BERT Tokenization} \\ 
 \hline
 irradiation & 'ir', '\#rad', '\#\#iation' \\ 
  \hline
 reactivity & 'react', '\#ivity'  \\
  \hline
 weld & 'w', \#eld \\
  \hline
 plenum & 'pl', '\#\#en', '\#\#um' \\
 \hline
 electrochemical & 'electro', '\#\#chemical' \\
  \hline
 carbide & 'car', '\#\#bid', '\#\#e' \\
  \hline
 boron & 'bo', '\#\#ron' \\
  \hline
 pellets & 'pe', '\#\#llet', '\#\#s' \\
  \hline
 oxides & 'oxide', '\#\#s' \\
  \hline
 annealing & 'anne', '\#\#aling' \\ 
 \hline
\end{tabular}
\caption{Table showing tokenization by BERT tokenizer of the BERT Custom Vocabulary - the vocabulary prepared by running sentence piece algorithm on NText. Note that BERT tokenizer often performs poorly on these samples.}
\label{table:1}
\end{table}

Next, we wanted to identify those words which BERT does not comprehend given its present vocabulary. First, Hugging Face's Pytorch implementation of BERT tokenizer is used to tokenize the Custom Vocab. The BERT tokenizer breaks the words into tokens, with each token being a part of BERT vocabulary. If a complete word is part of BERT vocabulary, it does not get broken into tokens. Otherwise, the word is tokenized, with each subword, except the first subword, being prefixed by \#\#. The tokenization of Custom-Vocab is shown by Table - \ref{table:1}. All those words which were retained as complete words were removed while the rest were stored in a CSV file with two columns: first column being the original word in Custom Vocab and second column being the tokenized form by BERT. Around 17 thousand words were tokenized by BERT indicating about 47\% overlap with BERT vocabulary. Intuitively, this process segregates those words which need special attention, as these words are a set of words which are considered important by Sentence Piece Library (and hence put in Custom-Vocab) and are not part of BERT Vocabulary.

With the help of domain experts, all the words were segregated into two groups: “good” and “bad”. “Good” meant that the tokenization is either successfully extracting the root word or can break into words which are close to their actual meaning (Figure - \ref{Courant_6}) and hence with pre-training on the domain-specific corpus, they will converge to their proper embedding. Bad meant that BERT is not even close to tokenizing these words and could create problems while performing downstream tasks (Figure - \ref{Courant_7})  

By manually iterating over "bad words", those words were clubbed which were having same/similar root words. For example: 'lubricant,' 'lubrication,' lubricated' were replaced by a single word 'lubric.' This approach is vital as it effectively increases the probability of seeing more data and hence would help in learning a more meaningful representation. In this way, 429 words were selected from the “bad” category. Bert Vocabulary contains about 1000 “UNUSED” tokens. 429 “UNUSED” tokens were replaced with the selected 429 words from “bad” category, hence forming NukeVOCAB.

% \begin{figure}[ht!] %!t
% \centering
% \includegraphics[width=3in, height = 3in]{fig6.png}
% \caption{List taken from “Good” Words.}
% \label{Courant_6}
% \end{figure}

\begin{table}
\centering
\large
\begin{tabular}{|c|c|} 
 \hline
 \textbf{"GOOD" Words} & \textbf{BERT Tokenization Split} \\ 
 \hline
 electrochemical & 'electro', '\#\#chemical' \\
  \hline
 conductivity & 'conduct', '\#\#ivity' \\
  \hline
 neutrons & 'neutron', '\#\#s' \\
  \hline
 ultrasonic & 'ultra', '\#\#sonic' \\
  \hline
 shutdown & 'shut', '\#\#done' \\
  \hline
 exchanger & 'exchange', '\#\#r' \\ 
 \hline
 polarisation & 'polar', '\#\#isation' \\
 \hline
 radiography & 'radio', '\#\#graphy' \\
 \hline
 warranty & 'warrant', '\#\#y' \\
 \hline
\end{tabular}
\caption{Table showing tokenization by BERT tokenizer of the "GOOD" Words. Note that BERT tokenizer performs reasonably well on these words, and thus shouldn't be added to NukeVocab.}
\label{table:2}
\end{table}

While performing the above operations, few notable observations were:

i) There were some non-English (Russian, German) words in the vocabulary, indicating that there were some portions of Russian and German texts in the corpus.

ii) Some words were combined without space like 'twodimensional'. These could be attributed to OCR errors, but as expected, BERT tokenizer was successfully able to split them, which makes BERT a suitable algorithm for domains like us having tough datasets.

\section{NQUAD: Nuclear Question Answering Dataset}

Question-Answering is a crucial part of human conversation \cite{15}, and hence the NukeBERT was decided to be evaluated on Question Answering task.  Unlike medical and general domain, the nuclear domain does not have any open source Question Answering Dataset. Such dataset is essential for the research community in this field to check the value of their models. Apart from checking the correctness, it is also useful for numerous applications. Hence it was decided to generate a new, high-quality Question Answering dataset.

For some time, we explored the idea of using the automatic question generation. We came across a few research papers (\cite{16}, \cite{17}, \cite{18}), which takes a paragraph as input and generates general questions. There are some rule-based methods, which take a paragraph and create questions.

The important reasons for deciding against using automatic question generation are:

i. Our dataset comprising research papers related to the nuclear domain turns out to be too difficult for the existing question generation systems. Due to the complex language, jargons, and complex sentences, the models were not able to generate meaningful and useful sentences.

% \begin{figure}[ht!] %!t
% \centering
% \includegraphics[width=3in, height = 3in]{fig7.png}
% \caption{ List taken from “bad” words. Notice the several words with root “lubric”.’ }
% \label{Courant_7}
% \end{figure}

\begin{table}
\centering
\large
\begin{tabular}{|c|c|} 
 \hline
 \textbf{"BAD" Words} & \textbf{BERT Tokenization Split} \\ 
 \hline
 lethargy & 'let', '\#\#har', '\#\#gy' \\
  \hline
 lubricant & 'lu', '\#\#bri', '\#\#can', '\#\#t' \\
  \hline
 lubricated & 'lu', '\#\#bri', '\#\#cated' \\
  \hline
 lubrication & 'lu', '\#\#bri', '\#\#cation' \\
  \hline
 luminescence & 'lu', '\#\#mine', '\#\#sc', '\#\#ence' \\
  \hline
 machining & 'mach', '\#\#ining' \\ 
 \hline
 ferritic & 'fe', '\#\#rri', '\#\#tic' \\
 \hline
 vapour & 'va', '\#\#pour' \\
 \hline
 flange & 'fl', '\#\#ange' \\
 \hline
\end{tabular}
\caption{Table showing tokenization by BERT tokenizer of the "BAD" Words. Note that BERT tokenizer performs very poorly on these words, and thus should be added to NukeVocab.}
\label{table:3}
\end{table}

ii. The question generation systems which use Named Entity Recognition (NER) \cite{19} to generate questions are generally ruled based which forms questions replacing for, e.g. nouns with a question having that noun as an answer. These types of questions doesn’t require semantic understanding and is rather mechanical.

The question-answering dataset is built with a format similar to SQuAD dataset. For that, research papers were randomly selected out of the 7000 research papers corpus. From these research papers, around 200 paragraphs were randomly selected to form the questions on.  Around 50 paragraphs were distributed to each domain expert, asking them to create questions on the paragraphs. They were encouraged to develop questions in their own words, the only restriction being that the answer must exactly lie within the paragraph as followed by SQuAD. The advised answer length was 8-10 words, though, in the final dataset, some accepted answers were longer than that also for generalization purposes. Experts found some paragraphs not suitable for forming questions, and hence, those were discarded. Finally, we were able to generate 612 questions on 181 paragraphs (Figure - \ref{Courant_8}).

The paragraph-questions-answers were recorded in shared google docs, responses from which were merged into an excel file. The file was randomly divided into two sections: a train set of 155 paragraphs, 536 questions, and dev set of 26 paragraphs, 76 questions.

To fine-tune the question-answering platform after training on SQuAD dataset, the custom-made Paragraph-Question-Answer dataset was converted to JSON with the structure similar to SQuAD v1. For turning it into SQuAD format, an already available platform for question generation was adapted for generating the NQuAD dataset. It is built using Angular as frontend with a spring boot backend using MongoDB database.

% \begin{figure}[ht!] %!t
% %\centering
% \includegraphics[width=3.5in, height = 3in]{fig8.png}
% \caption{Sample Paragraph-Questions-Answer triplets from NQuAD. Notice that the answers of the questions are strictly within the paragraphs similar to SQuAD.}
% \label{Courant_8}
% \end{figure}

\begin{table*}
\centering
\large
\begin{tabular}{|p{7.5cm}|p{5cm}|p{4cm}|} 
 \hline
 \textbf{Context Paragraph} & \textbf{Questions} & \textbf{Answers}\\ 
 \hline
Advanced Heavy Water Reactor (AHWR) fuel consists of (Th, Pu)O2 and (Th, 233U)O2. Thermochemistry of the compounds formed by the interaction of mixed oxide fuel and fission products is important in predicting the behaviour of this fuel in the reactor. Studies on the thermochemical properties of compounds in rare earth  tellurium–oxygen system are being carried out in our laboratory in order to predict the chemical behavior of the fission products. Cerium and tellurium are amongst the fission products which are formed during the burn up of the nuclear fuel with significant yield. Tellurium is corrosive in nature and can exist in various chemical states in irradiated fuel. It can cause embrittlement of clad. & 

What is the fuel used for Advanced Heavy Water Reactor (AHWR)? \newline \newline
What works are carried out in the laboratory in order to predict the chemical behavior of the fission products? \newline \newline \newline \newline \newline
What are formed during the burn up of the nuclear fuel with significant yield? &
  fuel consists of (Th, Pu)O2 and (Th, 233U) \newline \newline \newline
     Studies on the thermochemical properties of compounds in rare earth tellurium–oxygen system are being carried out in the laboratory.  \newline \newline
   Cerium and tellurium \\ 
 \hline
\end{tabular}
\caption{Table showing a sample paragraph and question answer pairs from NQuAD.}
\label{table:3}
\end{table*}

Since the dev set requires three answers per question, domain experts were asked to answer each of the 100 questions individually. The generated excel was again was converted into JSON format using the question generation platform described above.

\section{Experiment}

\subsection{BERT PRETRAINING}

Pre-Training BERT is a computationally expensive job and requires a large amount of data to pretrain. Our corpus of 8 Million words is two orders of magnitude less than the 3 Billion words corpus used by original BERT. It would be infeasible to pretrain BERT from scratch on this small corpus. Also, as mentioned earlier, there was almost 47\% overlap with the BERT vocab, which have already converged to proper representation in the pretrained weights. Pretraining from scratch would mean throwing away those learned embeddings. Pre-training from scratch becomes infeasible for domains like Nuclear, where the amount of data is pretty less. To tackle this issue, we used Nuclear Vocabulary (NukeVOCAB) in place of BERT vocabulary and pretrained it on the custom preprocessed corpus starting from BERT checkpoint. Due to Out of Memory issues, we used the batch size of 128 with maximum length set at 128. We pretrained it till the training loss more or less stopped decreasing, which happened after approximately 600000 steps. It took around 54 hours on a single Google Colab TPU for pre-training BERT. We will be referring this model as NukeBERT for further discussions.

\subsection{BERT Fine-Tuning}

For the base model, BERT was fine-tuned on SQuAD using the same configurations used by original BERT paper. On testing on NQuAD dev set, it scored 83.11 on exact match score and 92.66 on F1 score. The fine-tuning took around 8 hours on Google Cloud's Tesla k80 GPU (It is the version of GPU provided for free by google colab).

\begin{figure}
  \begin{subfigure}[b]{1.5in}
   \centering
    \includegraphics[width=1.6in, height = 3in]{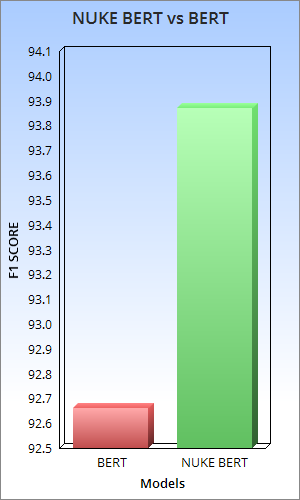}
    \caption{F1 scores }
    \label{fig:1}
  \end{subfigure}
  \begin{subfigure}[b]{2inin}
    \centering
    \includegraphics[width=1.6in, height=3in]{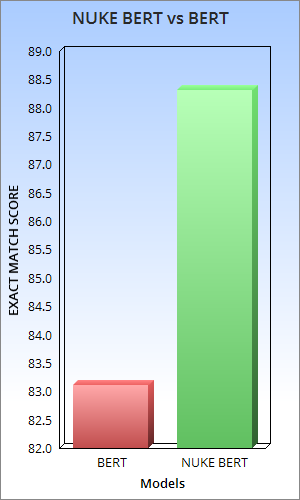}
    \caption{Exact Match}
    \label{fig:2}
  \end{subfigure}

\caption{Results of BERT and NukeBERT on NQuAD test-set}
\label{results}  
\end{figure}

With the same configuration as used by original BERT for fine-tuning, NukeBERT was fine-tuned on SQuAD dataset for two epochs. It took around 8 hours to train on Google Cloud GPU.

Further NukeBERT was fine-tuned on the NQuAD train set for three epochs at a learning rate of 3e-6.

\begin{table*}
\centering
\large
\begin{tabular}{|p{7cm}|p{5cm}|p{4.5cm}|} 
 \hline
 \textbf{Paragraph} & \textbf{Questions} & \textbf{Answer by NukeBERT}\\ 
 \hline
 
Walters and Cockraft (1972) might have been the first ones who used finite element method in analysing the creep of weldments.Coleman et al. (1985) incorporated a three-material model and used a parametric approach to cover a wide range of weld metal:base metal creep ratios. Ivarsson and Sandstrom (1980) studied the creep stress redistribution and rupture of welded
AISI 316 steel tubes by finite different method.Ivarsson (1983) studied the creep deformation of welded 12\% chromium steel tubes. Eggeler et al. (1994) performed a creep stress analysis of the welded pressure vessel made of modified 9Cr-1Mo material (P91) based on Norton’s creep law and specimen were studied using finite element by Segle et al. (1996) for the 1Cr-0.5Mo butt welded pipe.

&
 
Who used for the first time the finite element method in analysing the creep of weldments? \newline

Which method is used  in analysing the creep of weldments? \newline

Who used a three-material model for weld metal:base metal creep ratios ? \newline 

What does Sandstrom (1980) studied? \newline \newline

Who was associated with the finite different method? 

&

walters and cockraft \newline \newline \newline \newline

finite element method \newline \newline \newline \newline

coleman et al. \newline \newline

creep stress redistribution and rupture of welded aisi 316 steel tubes \newline

ivarsson and sandstrom \\

 \hline
\end{tabular}
\caption{Table showing qualitative performance of NukeBERT on unseen nuclear language passage.}
\label{table:5}
\end{table*}

\section{Results}

On testing, the NukeBERT achieved F1 score of 93.87 and exact match score of 88.31. Hence the model was able to achieve 5.21 improvement on exact match criteria and 1.22 improvement on F1 score (Figure - \ref{results}).

% ==================
% # IV. CONCLUSION #
% ==================

\section{Conclusion}
BERT requires large data for pre-training and generating meaningful word embeddings. However, the availability of data becomes a bottleneck for domains like nuclear whose data is profoundly different from world language. Hence, for areas like Nuclear energy, it becomes essential to use methods which we used for NukeVOCAB generation to optimize the use of meagre data. The improvement in F1 and exact match score validates the NukeBERT embeddings, which can be used for other downstream tasks too like sentence classification, named entity recognition among many NLP tasks. NukeBERT can be further generalized to many downstream tasks like named entity recognition, sentence segmentation among many, which could be highly important for nuclear industry. Moving further, the ideas used in this paper can be utilised with new transformers that have come up in recent research.  Data availability and creation in nuclear field, needs attention. This paper aims at providing momentum to research in nuclear domain.

% ==================
% # ACKNOLEDGMENTS #
% ==================

\section*{Acknowledgment}

We thank all colleagues of  Resource Management Group, IGCAR and BITS Pilani for their support and encouragement.

% ==============
% # REFERENCES #
% ==============

\bibliographystyle{IEEEtran}
\bibliography{IEEEabrv,biblio_traps_dynamics, refs}
% \def\bibfont{\large}
% \begin{thebibliography}{9}
% \def\bibfont{\large}

\end{document}